\documentclass[final]{cvpr}

\usepackage{times}
\usepackage{epsfig}
\usepackage{graphicx}
\usepackage{amsmath}
\usepackage{amssymb}

\usepackage{subcaption}
\usepackage{algorithm}
\usepackage{algorithmic}
\usepackage{gensymb}

\usepackage[pagebackref=true,breaklinks=true,colorlinks,bookmarks=false]{hyperref}

\newcommand\blfootnote[1]{%
  \begingroup
  \renewcommand\thefootnote{}\footnote{#1}%
  \addtocounter{footnote}{-1}%
  \endgroup
}

\def\cvprPaperID{3126} 
\def\confYear{CVPR 2021}

\begin{document}

\title{Attention-guided Image Compression by Deep Reconstruction of \\
          Compressive Sensed Saliency Skeleton}

\author{Xi Zhang\\
Shanghai Jiao Tong University\\
{\tt\small zhangxi\_19930818@sjtu.edu.cn}
\and
Xiaolin Wu$^\dagger$\\
McMaster Univeristy\\
{\tt\small xwu@ece.mcmaster.ca}
}


\maketitle
\thispagestyle{empty}

\begin{abstract}
   We propose a deep learning system for attention-guided dual-layer image compression (AGDL).
   In the AGDL compression system, an image is encoded into two layers, a base layer and an attention-guided refinement layer.
   Unlike the existing ROI image compression methods that spend an extra bit budget equally on all pixels in ROI, AGDL employs a CNN module to predict those pixels on and near a saliency sketch within ROI that are critical to perceptual quality.  Only the critical pixels are further sampled by compressive sensing (CS) to form a very compact refinement layer. Another novel CNN method is developed to jointly decode the two compression layers for a much refined reconstruction, while strictly satisfying the transmitted CS constraints on perceptually critical pixels.
   Extensive experiments demonstrate that the proposed AGDL system advances the state of the art in perception-aware image compression.

\end{abstract}
\blfootnote{$\dagger$ Corresponding author.}

\section{Introduction}

After decades of intensive research and development, visual signal compression techniques are approaching the rate-distortion performance limits. Any further significant improvements of bandwidth economy in visual communications have to come from smart human vision driven representations. In this direction the methodology of region-of-interest (ROI) image compression emerged about twenty years ago \cite{jpeg2000roi, nister1998lossless,atsumi1998lossy}. ROI compression is to exploit a well-known property of human vision: a viewer’s attention is not evenly distributed in all parts of an image. Instead, our attentions focus on one or few regions of greater interests than the rest of the image, which pertain to salient foreground object(s).
Background regions are delegated to our peripheral vision and hence have much lesser acuity. Playing this tapering of visual acuity away from ROIs, a ROI image compression method allocates a much lower bit budget to encode pixels outside of ROIs than those inside, and saves a significant number of bits without materially sacrificing visual quality of compressed images.

In this work, we sharpen the existing tool of ROI image compression and propose a deep learning system of attention-guided dual-layer image compression (AGDL).  In AGDL image compression, an image is encoded into two layers, a base layer $I_b$ and an attention-guided refinement layer $I_r$.  The base layer $I_b$ is a conventional compressed image of low bit rate (high compression), such as those produced by JPEG, JPEG 2000, WebP, BPG, etc.  The clarity of the base layer image just suffices to match the reduced level of acuity of peripheral vision.  It is up to the additional attention-guided refinement layer $I_r$ to boost the perceptual quality of ROI(s).


In existing ROI image compression methods, an extra bit budget is allocated to ROI and it is shared equally by all pixels in ROI.  But on a second reflection, we should be more discriminating than ROI and spend extra bits only on pixels that can contribute to perceptual quality after being refined. Instead of a contiguous region of interest, we introduce a much sparser representation called saliency sketch to highlight semantically significant structures within ROI. One step further, we define a so-called critical pixel set that is the intersection of the saliency sketch and the set of pixels that have large reconstruction errors. The critical pixel set specifies a skeletal sub-image that needs to be further sampled and refined.  For the saliency-driven refinement task, we take a more proactive approach than the straightforward CNN removal of compression artifacts
\cite{data,building,dncnn,dual,guo,galteri,dmcnn,idcn,neardcc,neartip,dcsc,stream,qgac}.
In the AGDL system design, the encoder takes and transmits $K$ additional samples of the critical pixel set in the form of compressive sensing (CS). The CS sampling produces novel critical information for the refinement layer, while having a very compact encoding of the novel information thanks to the small size of the critical pixel set.

The proposed AGDL image compression system needs to solve two key technical problems: 1. Detecting the saliency sketch and the critical pixels; 2. Refining the base layer with the CS measurements of the critical pixel set. The main technical contributions of this paper, besides the AGDL methodology, are the CNN solutions to the above two problems, one recognition and the other restoration.

\section{Related Works}

\subsection{End-to-end optimized image compression}
Toderici \etal \cite{toderici2015} exploited recurrent neural networks for learned image compression.
Some works \cite{balle2016, theis2017,agustsson2017} are proposed to approximate the non-differential quantization by a differentiable process to make the network end-to-end trainable.
Toderici \etal \cite{toderici2017} used recurrent neural networks (RNNs) to compress the residual information recursively.
Rippel \etal \cite{rippel2017,agustsson2019} proposed to learn the distribution of images using adversarial training to achieve better perceptual quality at extremely low bit rate.
Li \etal \cite{li2018} developed a method to allocate the content-aware bit rate under the guidance of a content-weighted importance map.
\cite{mentzer2018,balle2018,minnen2018,lee2018,nonlinear} focused on investigating the adaptive context model for entropy estimation to achieve a better trade-off between reconstruction errors and required bits (entropy), among which
the CNN methods of \cite{minnen2018,lee2018} are the first to outperform BPG in PSNR.

\subsection{ROI image compression}
In the AGDL image compression system outlined above, the first step is to understand the image semantic composition and segment salient foreground objects.  Detecting salient objects is a research topic in its own right. Recently, good progress has been made on this topic thanks to advances of deep learning research in computer vision, with a number of CNN segmentation methods published to extract salient objects from the background \cite{
picanet,progressive,detect,bidirectional,clutter,contour,basnet,simple,cascaded,attentive,pyramid,egnet,employing,high,deeper,weakly,decoupling,multi};
they can be applied to compute ROIs. But for the purpose of image compression, we need to push further and seek for the shortest description of salient objects.

ROI based image compression, which is less discriminative than AGDL in selecting critical pixels for refinement, was an active research topic at the time of JPEG 2000 standard development \cite{jpeg2000roi,nister1998lossless,atsumi1998lossy}.  Unlike JPEG that uses block DCT of very low spatial resolution ($8 \times 8$ superpixel), JPEG 2000 is a two-dimensional wavelet representation and it can operate on images in good spatial resolution. This property makes ROI image compression possible. In conventional ROI coding, extra bits are spent to encode the ROI segment. As the ROI shape is determined by the contours of foreground objects, a flexible spatial descriptor inevitably consumes a significant amount of extra bandwidth. This cost of side information on ROI geometry could offset any rate-distortion performance gain made by ROI compression. This dilemma can be overcome by deep learning, as we demonstrate in the subsequent development of AGDL system and methods. By training a CNN to satisfactorily predict the saliency skeleton within ROI, AGDL compression strategy can enjoy the benefits of attention-guided compression free of side information.

Very recently, a CNN based ROI image compression method was published \cite{SJTUTIP}. This is a pure CNN compression system of the standard auto-encoder architecture. The authors proposed the idea of extracting some CNN features specifically for the ROI. As explained in the introduction, the saliency sketch of AGDL is far more discriminative than a contiguous ROI; therefore it leads to more efficient use of extra refinement bits.  Furthermore, we use CS measurements of critical pixels to exert input-specific constraints on the solution space of the underlying inverse problem, rather than solely relying on the statistics of the training set as in \cite{SJTUTIP}.  Finally, there is a drastic difference in encoder throughput between the method of \cite{SJTUTIP} and our method.  The base layer encoder of the proposed AGDL system can be any conventional image compressor (e.g. JPEG, JPEG 2000, WebP, BPG, etc.), which has a complexity orders of magnitude lower than CNN auto-encoder.







\section{AGDL Compression System}

In this section, we will introduce the design of the proposed AGDL image compression system,
and two key technical contributions:
1. detecting saliency sketch and critical pixel set from the compressed base layer image;
2. Refining the base layer image with the CS measurements of the critical pixel set.

\subsection{Overview}

\begin{figure*}[t]
\centering
\includegraphics[width=0.95\textwidth]{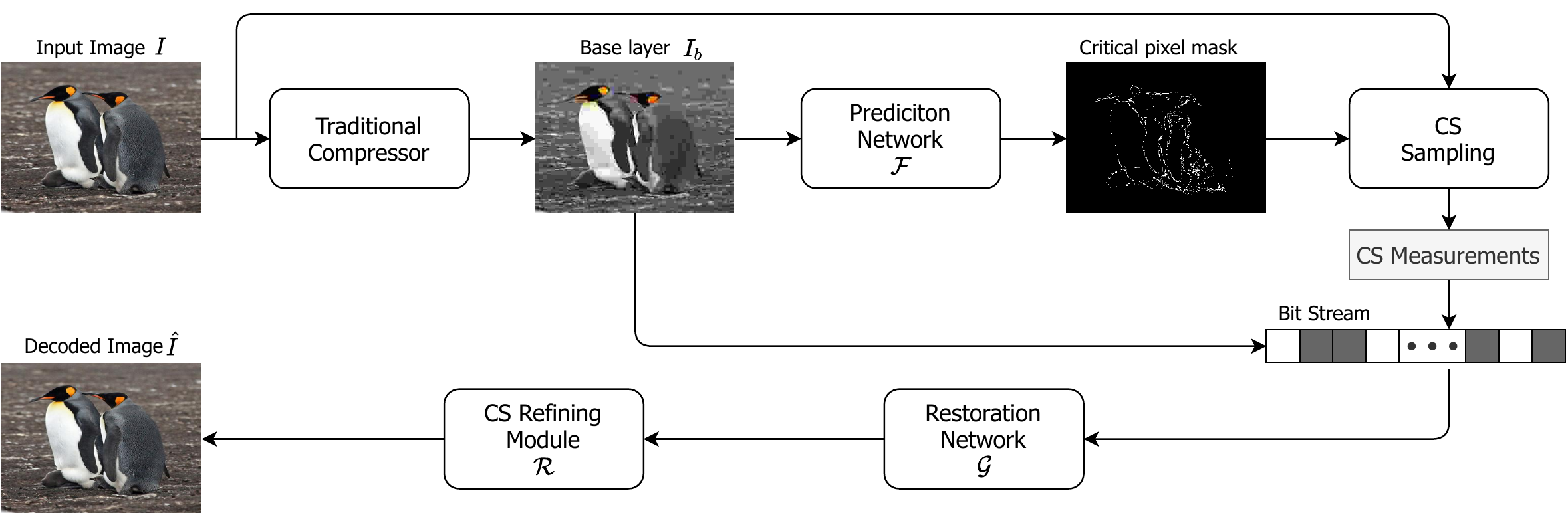}
\caption{The overall framework of the proposed AGDL image compression system.}
\label{agdl}
\end{figure*}

The overall framework of the proposed AGDL image compression system is shown in Fig.~\ref{agdl}.
It consists of a two-stage encoder and a joint decoder.
Given an image $I$ to be compressed, AGDL compression system first encodes $I$ to a base layer $I_b$ using a traditional image compressor, and then predicts the critical pixel mask $\mathbb{C}$ from the base layer $I_b$ using a deep neural network $\mathcal{F}$.  The resulting critical pixel mask $\mathbb{C}$ is used to extract the set of critical pixels $\mathbf{c}$.
After that, AGDL system performs compressive sensing (CS) on the detected critical pixel set and transmits the CS measurements $\mathbf{y}$ along with the base layer $I_b$.
The decoder takes the base layer $I_b$ and the CS measurements $\mathbf{y}$ of the critical pixel set as input to produce a refined image $\hat{I}$ with highlighted semantic structures by a restoration network $\mathcal{G}$ and a CS refining module $\mathcal{R}$.

\subsection{Saliency sketch and critical pixels}

\begin{figure}[t]
\centering
	\begin{subfigure}{0.45\linewidth}
		\centering
		\includegraphics[width=0.98\linewidth, height=0.7\linewidth]{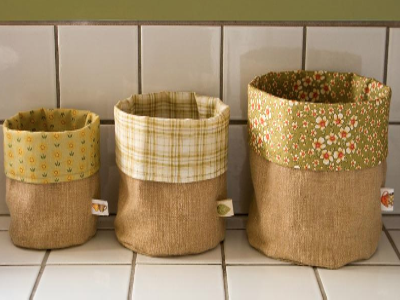}
		\caption{Image}
		\label{img}		
	\end{subfigure}
	\begin{subfigure}{0.45\linewidth}
		\centering
		\includegraphics[width=0.98\linewidth, height=0.7\linewidth]{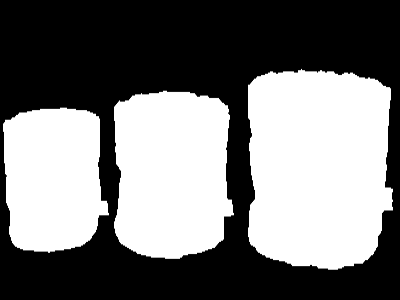}
		\caption{ROI}
		\label{roi}		
	\end{subfigure}
	\begin{subfigure}{0.45\linewidth}
		\centering
		\includegraphics[width=0.98\linewidth, height=0.7\linewidth]{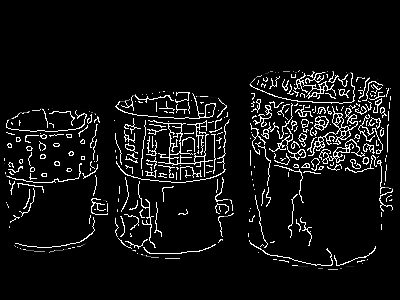}
		\caption{Saliency sketch}\label{ske}		
	\end{subfigure}
	\begin{subfigure}{0.45\linewidth}
		\centering
		\includegraphics[width=0.98\linewidth, height=0.7\linewidth]{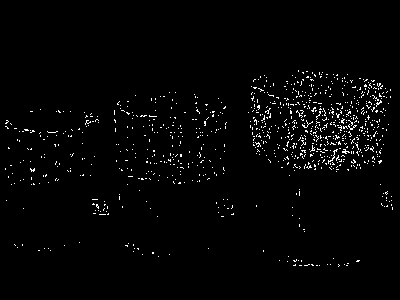}
		\caption{Critical pixel mask}\label{cri}		
	\end{subfigure}
\caption{Examples of the natural image, ROI map, and the proposed saliency sketch and critical pixel mask.}
\label{}
\end{figure}

Existing ROI image compression methods, including the recently proposed pure CNN ROI compression system~\cite{SJTUTIP}, weigh all pixels in ROI equally.
However, not all pixels in ROI carry the same significance to visual quality.
For example in Fig~\ref{roi}, the featureless power portions of the three baskets matter much less to visual perception than the textured upper portions.
A rate-distortion more efficient way of coding is to allocate more bits only to pixel structures that contribute the most to improving perceptual quality, such as edges and textures.
To this end, we introduce a much sparser presentation than ROI, called saliency sketch, which is defined as the edge map of the object(s) in ROI, as shown in Fig.~\ref{ske}.

In fact, we can be even more selective than saliency sketch, if considering the recent progresses made on CNN based compression artifact removal (CAR) techniques~\cite{dncnn,dual,dmcnn,qgac}.  These learning methods can restore many pixels belonging to saliency sketch, and the CNN recoverable pixels need not be additionally sampled and transmitted.  Thus the AGDL encoder only needs to send new information on the pixels that belong to saliency sketch and but also have large reconstruction error.  We define these pixels critical pixels.

Denoting the edge skeleton of $I$ by $\Omega_s$, the ROI of $I$ by $\Omega_i$, the set of pixels of large reconstruction errors after CAR by $\Omega_e$, then the critical pixel mask $\mathbb{C}$ can be represented as:
\begin{align}
\centering
\mathbb{C} = \Omega_s \cap \Omega_i \cap \Omega_e
\label{mathbbc}
\end{align}
In Fig.~\ref{cri}, the critical pixel mask indicates the locations of the critical pixels.
The critical pixel set specifies a skeletal sub-image that needs to be further sampled and refined.

\subsection{Detecting critical pixel set}

In traditional ROI image coding, the ROI geometry is explicitly encoded and therefore is a part of compression code stream.  The extra bits required to transmit the ROI shape could offset any rate-distortion performance gain made by ROI compression.  This dilemma can be overcome by deep learning if a CNN can learn to predict the ROI mask from the base layer image $I_b$.  This eliminates the need to transmit the ROI shape because the decoder can make the same ROI prediction as the encoder.

In the AGDL image compression system, we push further and drive a CNN $\mathcal{F}$ to predict the critical pixel mask $\mathbb{C}$ that is a subset of ROI from the base layer $I_b$.  This learning task is more demanding, but it is nevertheless feasible because the critical pixel mask $\mathbb{C}$ of an image can be computed to generate paired data for supervised learning.  This is a strategy of squeezing out coding gains by computation power and big data.

Specifically, we adopt an existing CAR network called DnCNN~\cite{dncnn} to initially restore base layer $I_b$ and then identify the set $\Omega_e$ of those pixels that still have large restoration errors.  In addition, we use a salient object network BASNet~\cite{basnet} to calculate the ROI region $\Omega_i$, and detect the edge skeleton $\Omega_s$ using Canny operator.

Given $\Omega_i$, $\Omega_s$ and $\Omega_e$, the critical pixel mask $\mathbb{C}$ is determined.
So we can build paired data (baser layer images $I_b$ and the corresponding critical pixel masks $\mathbb{C}$) to train the prediction network $\mathcal{F}$.
Let $\mathcal{F}$ be the prediction network:
\begin{align}
\centering
\mathbb{C} = \mathcal{F}(I_b)
\end{align}

\begin{figure}[t]
	\centering
	\includegraphics[width=1.0\linewidth]{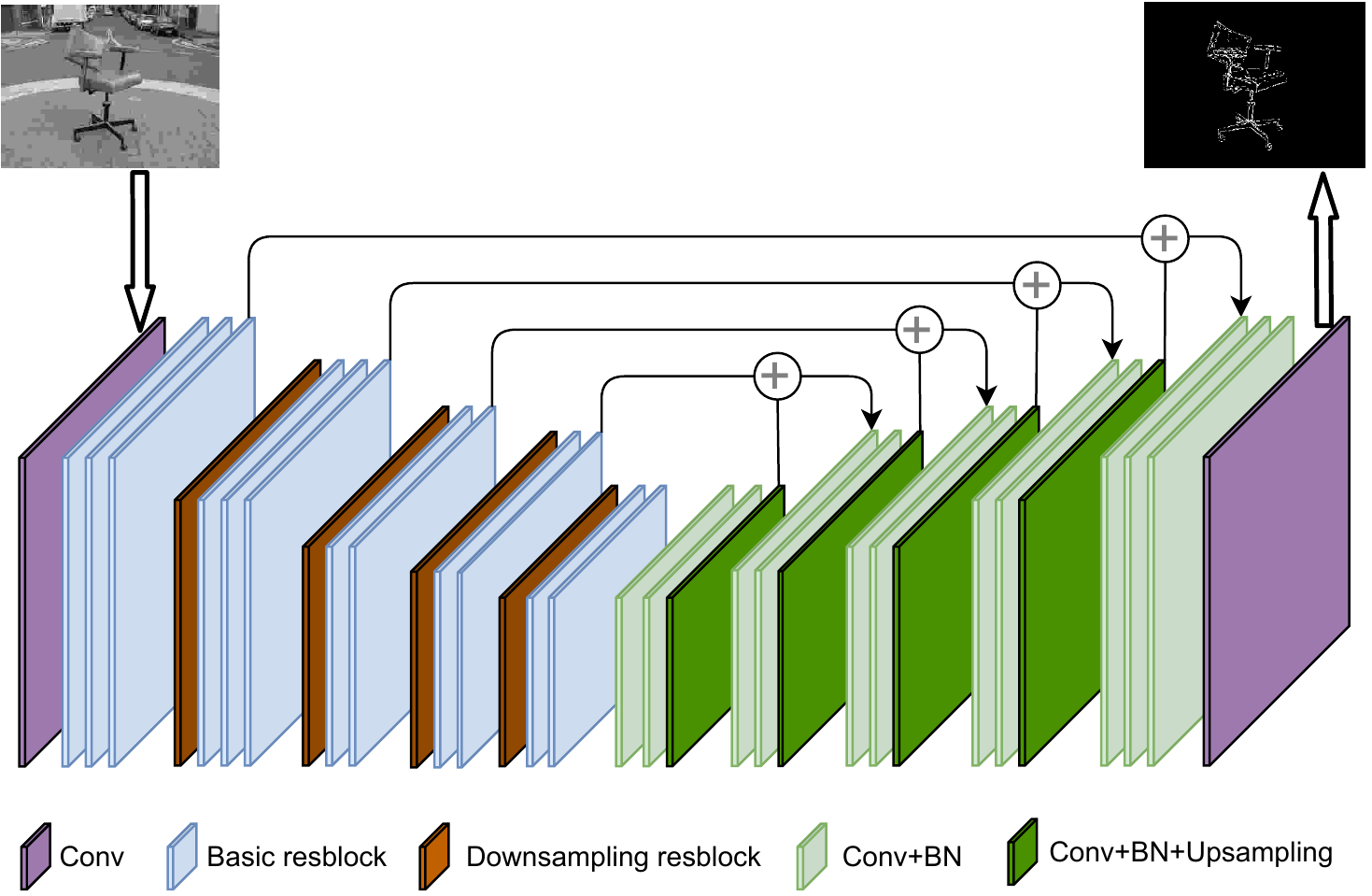}
	\caption{Architecture of the proposed critical pixel mask prediction network $\mathcal{F}$.}
	\label{predF}		
\end{figure}

The architecture of the proposed prediction network $\mathcal{F}$ is revised from BASNet~\cite{basnet}, a network designed for salient object detection.
As shown in Fig.~\ref{predF}, the prediction network $\mathcal{F}$ is a U-Net-like Encoder-Decoder network~\cite{unet}, which learns to predict critical pixel mask from base layer image.
We design the critical pixel mask prediction network as an Encoder-Decoder architecture because it is able to capture high level global contexts and low level details at the same time~\cite{unet,fcn}.
The encoder part has an input convolution layer and five stages comprised of residual blocks.
The input layer has 64 convolution filters with size of 3$\times$3 and stride of 1.
The first stage is size-invariant and the other four stages gradually reduce the feature map resolution by downsampling resblocks to obtain a larger receptive field.
The decoder is almost symmetrical to the encoder. Each stage consists of three convolution layers followed by a batch normalization and a ReLU activation function. The input of each layer is the concatenated feature maps of the up-sampled output from its previous layer and its corresponding layer in the encoder.

The critical pixel set $\mathbf{c}$ can be extracted based on the predicted critical pixel mask $\mathbb{C}$, and then rearranged into a column vector. After that, AGDL compression system performs Compressed sensing on the critical pixel set $\mathbf{c}$ with a full row rank, fat CS sampling matrix $H$ (far fewer rows than columns):
\begin{align}
\centering
\mathbf{y} = H \cdot \mathbf{c}
\end{align}
where $\mathbf{y}$ is the CS measurements of the critical pixel set. The CS measurements $\mathbf{y}$ and base layer $I_b$ will be transmitted to the decoder end.

\subsection{Dual-layer joint decoding}

The most important and technically involved component of the AGDL image compression system is its CNN decoder.  The task of AGDL decoding is to refine the JPEG-coded base layer $I_b$, aided by the CS-coded side information on saliency skeleton.
Specifically, the AGDL decoder receives the base layer $I_b$ and refinement layer $I_r$ (CS measurements of critical pixels), and then jointly decodes the two layers to produce a refined image $\hat{I}$ which strictly satisfies the CS constraints.  In essence, the AGDL decoder is a heavy-duty CNN that removes the compression artifacts of the base layer image $I_b$ with encoder-supplied strong priors on ROI.

By satisfying the CS constraints we mean that after the critical pixel set $\hat{c}$ in the CNN refined image $\hat{I}$ is sampled by the CS sampling matrix $H$, the resulting CS measurements $\hat{y}$ equal to the received measurements $\mathbf{y}$, that is
\begin{align}
\centering
H \cdot \hat{\mathbf{c}} = \mathbf{y}
\end{align}
To the best of our knowledge, we are the first to impose such constraints on CNN outputs. This way of confining the solution space of an inverse problem in CNNs poses a technical challenge.  We overcome the difficulty by cascading a restoration network $\mathcal{G}$ and a CS refining module $\mathcal{R}$, in which the latter constrains the output of the former by the CS measurements.  The joint decoding process can be formulate as:
\begin{align}
& I_g = \mathcal{G}(I_b) \\
& \hat{I} = \mathcal{R}(I_g, \mathbf{y})
\end{align}
where $I_b$ is the decoded result of a traditional image compressor (decompressed image); the restoration network $\mathcal{G}$ performs a post-processing on $I_b$, called soft decoding, aiming to remove compression artifacts in $I_b$.  The result of soft decoding is a restored image $I_g$.
The final step of the AGDL system is to adjust the set of critical pixels in $I_g$, denoted by $\mathbf{c}_g$, so that their values strictly satisfy the set of $K$ CS measurements.  Among all possible such $K$-dimensional adjustment vectors $\delta$, the one $\delta_*$ of the minimum $\ell_2$ norm generates the final refinement image $\hat{I} = I_g + \delta_*$.


Next we develop the CS refining module $\mathcal{R}$ that imposes constraints on the final output of the AGDL system.
Firstly, $I_g$ must not satisfy the CS constraint, that is
\begin{align}
\centering
H \cdot \mathbf{c}_g \neq \mathbf{y}
\end{align}
where $\mathbf{c}_g$ is the critical pixel set in the restored image $I_g$.
We hope to make the minimum adjustment to the output image $I_g$ (or to the critical pixel set $\mathbf{c}_g$) so that the adjusted image can satisfy the CS constraint. This forms the following optimization problem:
\begin{align}
\centering
\text{minimize} \quad & ||\mathbf{\delta}|| \\
\text{subject to} \quad & H \cdot (\mathbf{c}_g + \delta) = \mathbf{y}
\label{opt}
\end{align}

Since the CS sampling matrix $H$ is full row rank, so the above optimization problem has the solution that is:
\begin{align}
\centering
\delta^* = H^T(HH^T)^{-1} \cdot (\mathbf{y} - H \cdot \mathbf{c}_g)
\label{sol}
\end{align}
This is the classical least-norm solution of undetermined equations. Detailed solving steps will be given in the supplementary material.
Let $\hat{\mathbf{c}} = \mathbf{c}_g + \delta^*$, so the adjusted critical pixel set $\hat{\mathbf{c}}$ satisfies the CS constraint.  It is noteworthy that the adjustment is linear, so it can participate in the back propagation.

In the design of the restoration network $\mathcal{G}$, we adopt a dual-domain (pixel domain and transform domain) network to take full advantage of redundancies in both pixel and transform domains~\cite{dmcnn,davd}. In most traditional image compression methods, images are converted to a transform domain (e.g., DCT, wavelet, etc.) and then quantized. The encoder prior information contained in the transform domain can help improve the performance of soft-decoding.

\begin{figure}[t]
	\centering
	\includegraphics[width=1.0\linewidth]{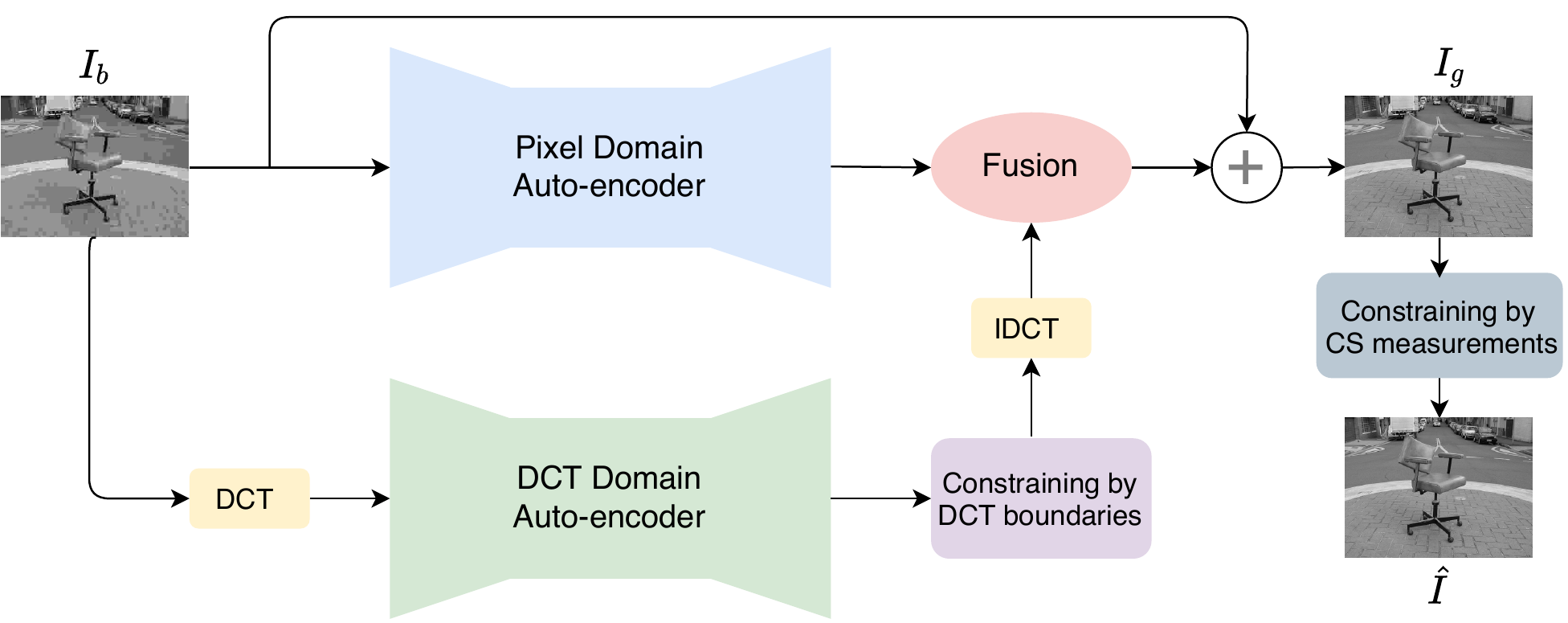}
	\caption{Architecture of the decoder, including the restoration network $\mathcal{G}$ and the CS refining module $\mathcal{R}$.}
	\label{restG}		
\end{figure}

The base layer of the AGDL system can be any of existing image compression methods.  In this paper, we choose JPEG as the base layer to develop the restoration network $\mathcal{G}$, as it is the most common compression method.
As shown in Fig.~\ref{restG}, the proposed restoration network $\mathcal{G}$ has two branches, one operating in pixel-domain and the other in DCT domain.
The pixel-domain branch is to restore the pixel values directly, while the DCT-domain branch aims to recover the DCT coefficients of the ground truth. 
The fusion network combines these two branches to produce
the restored image $I_g$. After CS refinement, $\hat{I}$ is used to calculate loss to optimize the network $
\mathcal{G}$.

Now we are at the point to present the overall pipeline of AGDL compression system in Algorithm.~\ref{alg_agdl}.
\begin{algorithm}[!t]
\caption{ Framework of AGDL compression system.}
\label{alg_agdl}
\hspace*{0.02in} {\bf Input:}
The original image, $I$; \\
\hspace*{0.02in} {\bf Output:}
The decoded image, $\hat{I}$; \\
\hspace*{0.02in} {\bf Encoding:}
\begin{algorithmic}[1]
\STATE Encoding $I$ into a base layer $I_b$ using JPEG;
\STATE Predicting critical pixel mask $\mathbb{C}$ from the base layer $I_b$ by the prediction network $\mathcal{F}$, $\mathbb{C} = \mathcal{F}(I_b)$;
\STATE Extracting critical pixel set $\mathbf{c}$ based on $\mathbb{C}$;
\STATE Applying compressive sensing on $\mathbf{c}$, $\mathbf{y}=H \cdot \mathbf{c}$;
\STATE Transmitting $I_b$ and $\mathbf{y}$;
\end{algorithmic}
\hspace*{0.02in} {\bf Decoding:}
\begin{algorithmic}[1]
\STATE Soft-decoding $I_b$ by the network $\mathcal{G}$, $I_g = \mathcal{G}(I_b)$;
\STATE Calculating the minimum adjustment to satisfy the CS constraint, $\delta^* = H^T(HH^T)^{-1} \cdot (\mathbf{y} - H \cdot \mathbf{c}_g)$;
\STATE Applying the adjustment, $\hat{I} = I_g + \delta^*$;
\STATE Output the final refinement image $\hat{I}$;
\end{algorithmic}
\end{algorithm}


\section{Experiments}
In this section, we introduce the implementation details of the proposed AGDL image compression system.
To systematically evaluate and analyze the performance of the AGDL compression system, we conduct extensive experiments on two scenarios: portrait and general objects, and compare our results with several stat-of-the-art methods.

\begin{figure*}[t]
	\centering
	\includegraphics[width=0.46\linewidth]{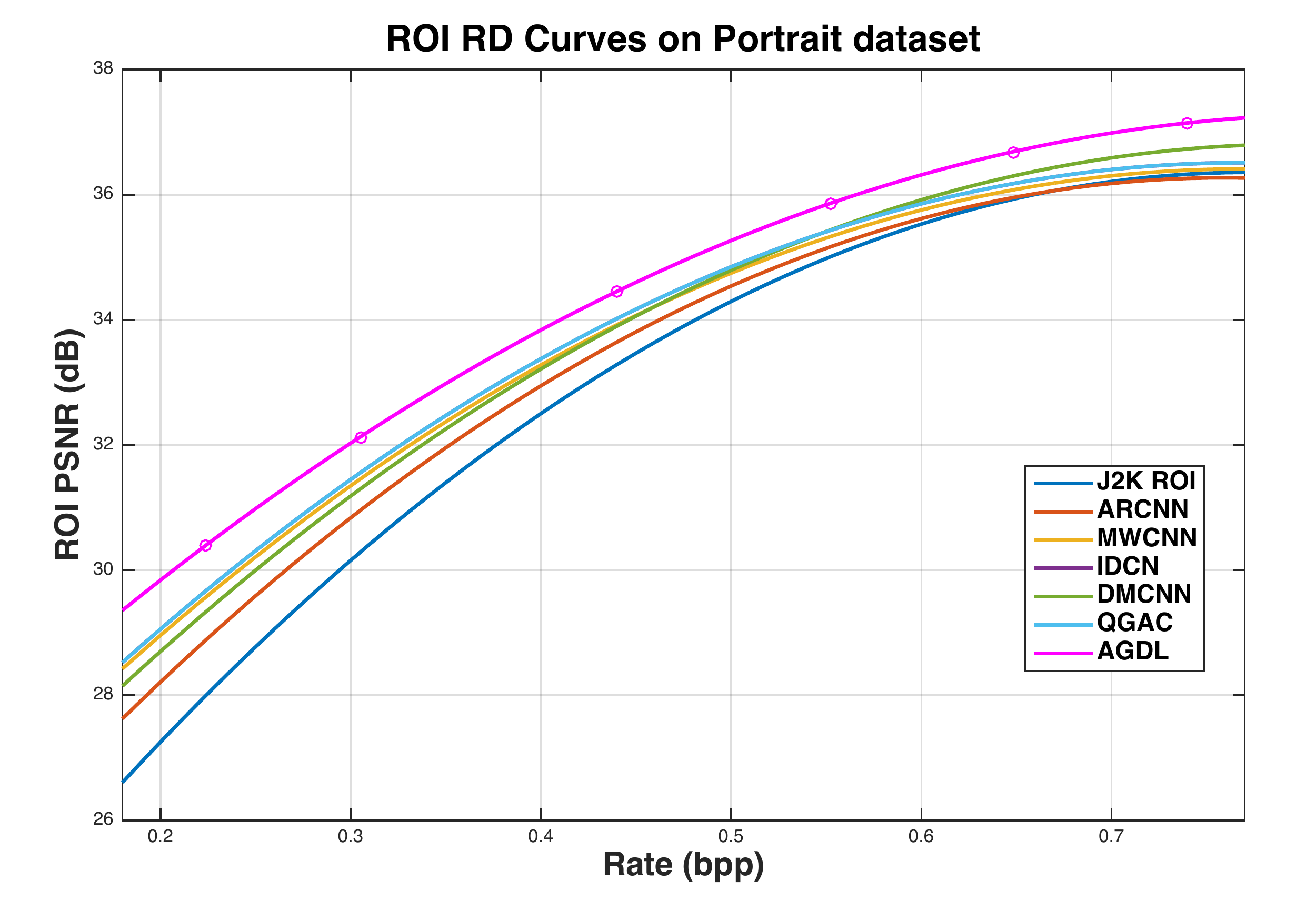}
	\hfill
	\includegraphics[width=0.46\linewidth]{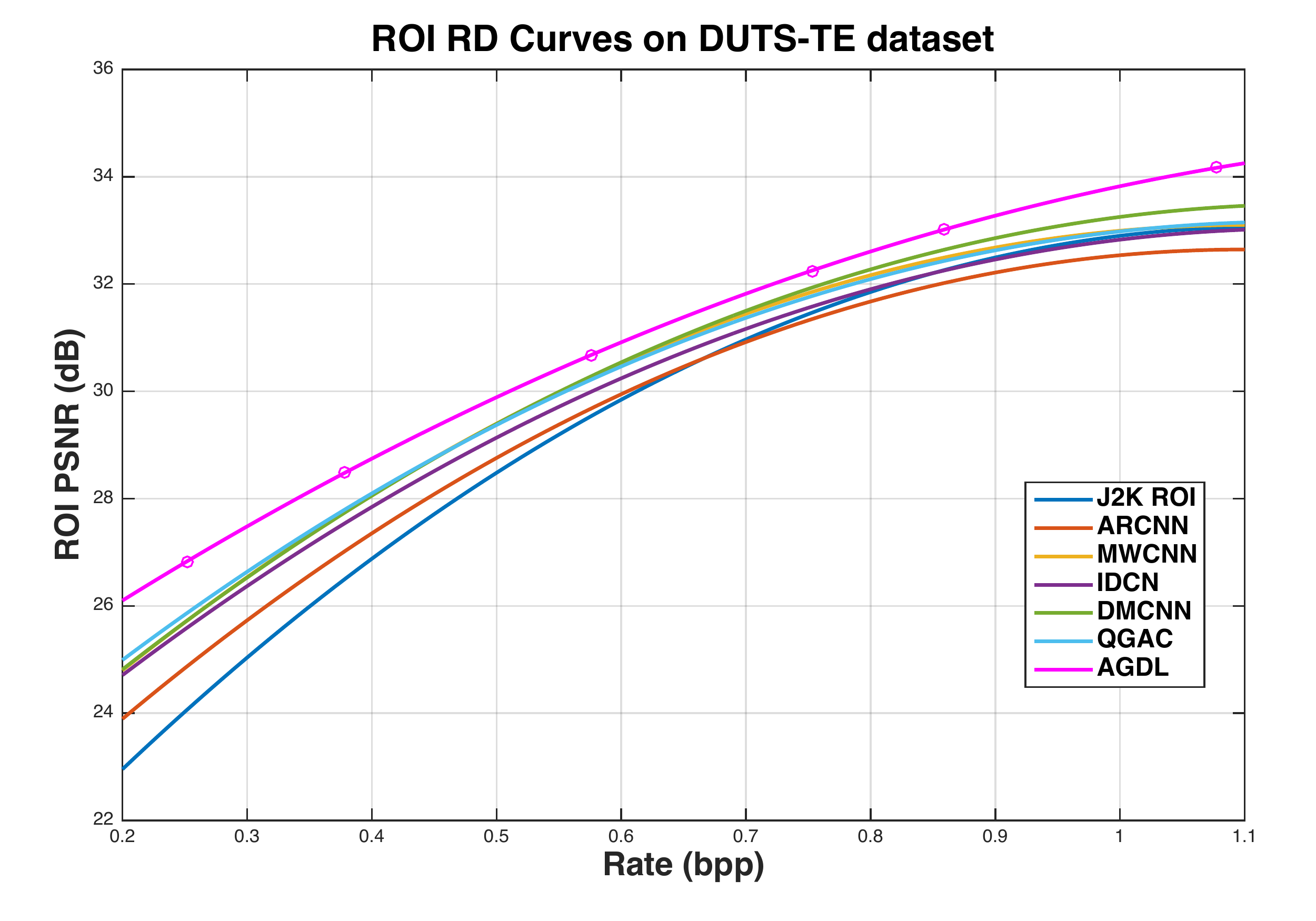}
	\caption{ROI RD curves of the competing methods on Portrait and DUTS-TE datasets.}
	\label{rd_curve}		
\end{figure*}

\subsection{Dataset}

\textbf{Portrait}.
We adopt the portrait dataset provided by Shen et al.~\cite{shen} for training and evaluation.
It contains 2000 images of 600 $\times$ 800 resolution where 1700 and 300 images are split as training and testing set respectively. To overcome the lack of training data, we augment images by utilizing rotation and left-right flip, as suggested in~\cite{seo}.  Each training image is rotated by $\left[ -15\degree, 15\degree \right]$ in steps of $5\degree$ and left-right flipped, which means that a total of 23800 training images are obtained.

\textbf{General objects}.
In the scenario for general objects, we adopt the DUTS dataset~\cite{duts} for training and testing. Currently, DUTS is the largest and most frequently used dataset for salient object detection.
DUTS dataset consists of two parts: DUTS-TR and DUTS-TE.
DUTS-TR contains 10553 images in total.  We augment this dataset by horizontal flipping to obtain 21106 training images.  DUTS-TE, which contains 5019 images, is selected as our evaluation dataset.

All these images are resized to 300 $\times$ 400 resolution for training and evaluation. We choose JPEG as the traditional image compressor of the AGDL system, as JPEG is the most widely used image compression method. For both scenarios, we compress the images using JPEG with quality factor in $[10, 100]$ in steps of 10 to form a multi-rate training set. All training and evaluation processes are performed on the luminance channel (in YCbCr color space).


\begin{figure*}[t]
	\centering
	\includegraphics[width=0.98\linewidth, height=1.02\linewidth]{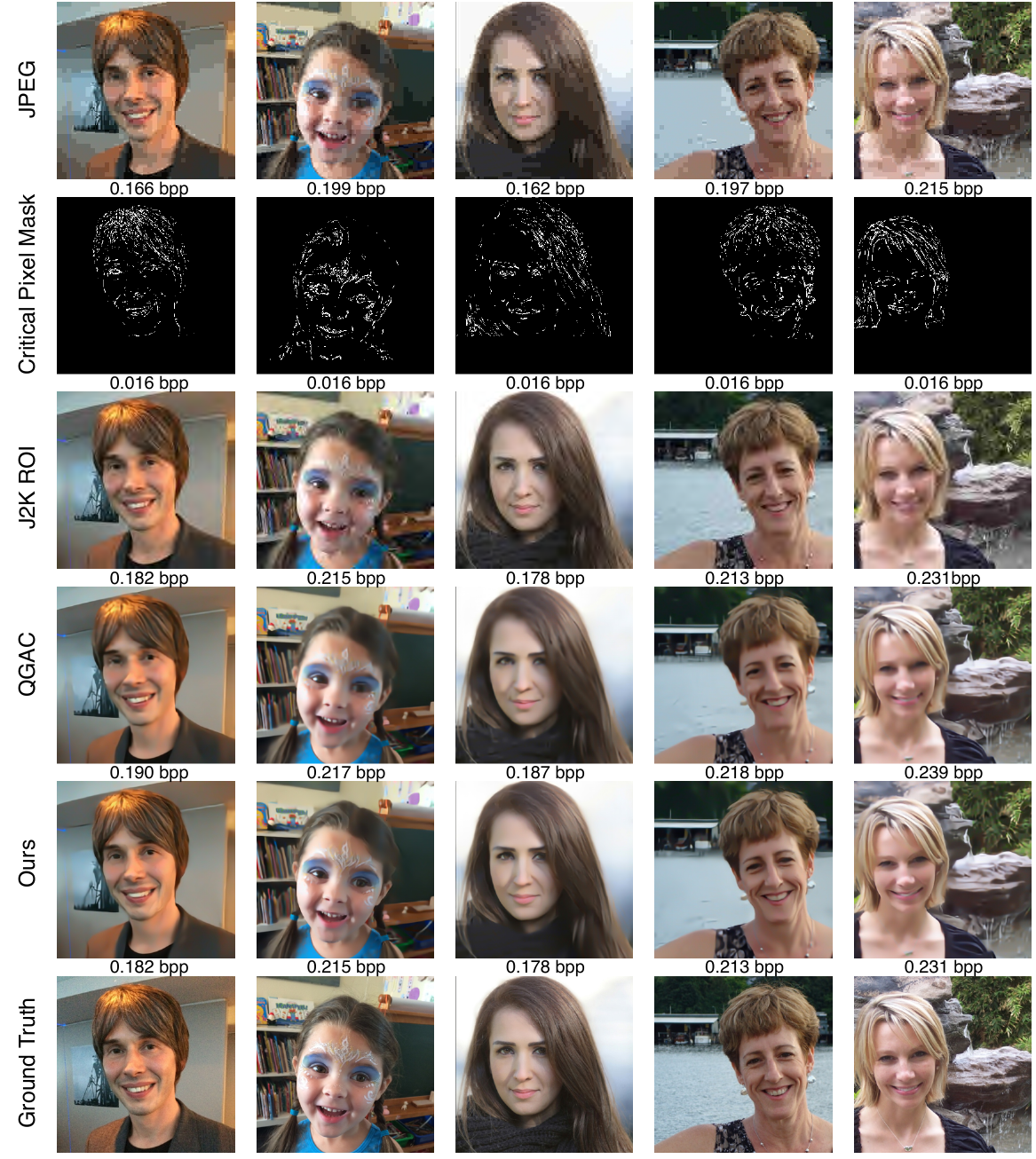}
	\caption{Visual comparisons of different methods on portraits.}
	\label{visual_protrait}		
\end{figure*}
\begin{figure*}[t]
	\centering
	\includegraphics[width=0.98\linewidth, height=1.0\linewidth]{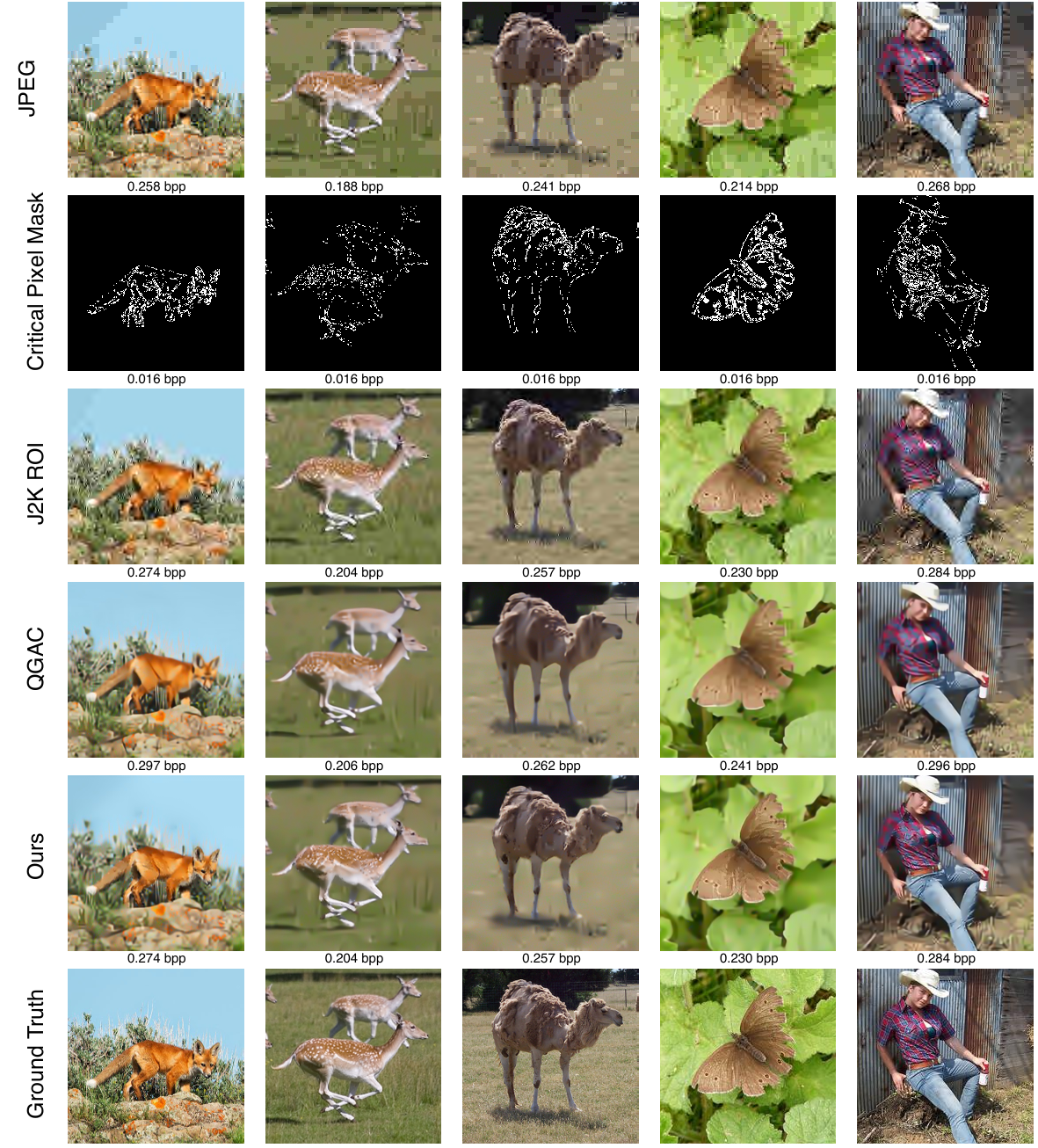}
	\caption{Visual comparisons of different methods on general objects.}
	\label{visual_duts}
\end{figure*}

\subsection{Training details}
Totally, we have two networks to train, a prediction network $\mathcal{F}$ and a restoration network $\mathcal{G}$. Next, we introduce the training details of the two networks separately.

\textbf{Prediction network $\mathcal{F}$}.
To train the network $\mathcal{F}$ for predicting critical pixel mask, we first adopt DnCNN~\cite{dncnn} to initially restore the JPEG-coded images and then identify the set $\Omega_e$ of those pixels that still have large restoration errors (error $> 8$). In addition, we use a salient object network BASNet~\cite{basnet} to calculate the ROI region $\Omega_i$, and detect the edge skeleton $\Omega_s$ using Canny operator. Then, we get the critical pixel mask $\mathbb{C}$ according to Eq.~\ref{mathbbc}.
The critical pixel mask $\mathbb{C}$ is a binary mask, in which $1$ means the current location is critical pixel and $0$ vice versa.
The prediction network $\mathcal{F}$ takes JPEG-coded images as input and outputs the corresponding critical pixel masks, so it solves a binary classification problem for each pixel location.
To this end, we train $\mathcal{F}$ using the Binary Cross Entropy (BCE) loss function.
When inferring, prediction network $\mathcal{F}$ outputs a probability value in $\left[0, 1\right]$ for each pixel location, indicating the probability of being a critical pixel. Top $K$ pixels ranked by probability form the critical pixel set to be further sampled and transmitted.
More details about the CS sampling matrix $H$ are given in the supplementary material.

\textbf{Restoration network $\mathcal{G}$}.
To reduce the risk of over-fitting, the restoration network $\mathcal{G}$ is pretrained using the DIV2K~\cite{edsr} and Flikr2K~\cite{edsr} datasets.
After pretrained, the restoration network $\mathcal{G}$ is fine-tuned on portrait dataset~\cite{shen} and DUTS-TR~\cite{duts} separately, under the constraints of the CS measurements.
$\mathcal{L}_1$ loss is adopted to optimize the restoration network $\mathcal{G}$.


All training processes use the Adam~\cite{adam} optimizer by setting $\beta_1 = 0.9$ and $\beta_2 = 0.999$, with a batch size of 16. The network is trained with 100 epochs at the learning rate of $10^{-4}$ and other epochs with learning rate of $10^{-5}$.
The algorithms are implemented in the MindSpore framework.

\subsection{Comparison with state-of-the-art methods}
To demonstrate the advantages of the proposed AGDL compression system, we compare AGDL with several other compression systems, in which JPEG is also used as the compressor and several deep-learning based compression artifact reduction methods
ARCNN~\cite{arcnn}, MWCNN~\cite{mwcnn}, IDCN~\cite{idcn}, DMCNN~\cite{dmcnn}, QGAC~\cite{qgac} are used as the soft decoder.
In order to factor out the effects of different training sets and conduct a fair comparison, we fine-tune all CNN networks in the comparison group using the same datasets (Portrait and DUTS) in our experiments.
We also compare AGDL with JPEG2000's ROI coding which is implemented in Kakadu JPEG2000 software.
In the AGDL system, the total bit rates need to be transmitted are the sum of the rates of JPEG-coded base layer and the CS-coded side information. To facilitate fair rate-distortion performance evaluations, for each test image, the rates of the competing compression systems are adjusted to match or be slightly higher than that of the AGDL compression system.

\textbf{Quantitative results.}
We present rate distortion (RD) curves of ROI in Fig.~\ref{rd_curve}. The rate is calculated by
bits consumed to encode the entire image averaged per pixel (bpp), and the distortion is measured by the PSNR of the ROI area. For AGDL, the rate is the sum of the bits consumed by the JPEG-coded base layer and the CS-coded side information.
As shown in Fig.~\ref{rd_curve}, the proposed AGDL compression system outperforms all the competing methods by a large margin, on both portrait images and general object images.
For portrait images, the PSNR gain obtained by AGDL is relatively uniform in bit rate.
However, for general objects, the PSNR gain is unevenly distributed. Specifically, the more extreme the bit rate, the greater the PSNR gain.

\textbf{Qualitative results.}
In addition to the quantitative results of RD curves, we also present the visual comparisons of different methods, as shown in Fig.~\ref{visual_protrait} and \ref{visual_duts}. QGAC~\cite{qgac} is the state-of-the-art CNN method for compression artifacts reduction, so we only show QGAC's results for visual comparison due to page limit. The complete visual comparisons of all competing methods will be given in the supplementary material. In the visual comparisons, we add the color channels (CbCr) back for the best visual quality.  In Fig.~\ref{visual_protrait}, we can see that the AGDL compression system can preserve facial features better than the state-of-the-art QGAC method and J2K ROI compression (note clearer eyes and hair, sharper muscle contours).
For general objects, Fig.~\ref{visual_duts} shows us that the AGDL system is able to preserve the small structures with the help of CS constraints, such as the spots on the sika deer and the lines on the butterfly.  In addition, AGDL can make animal hair more realistic, while QGAC makes the hair look too smooth.


\section{Conclusion}
\vspace{-0.1cm}
We present a deep learning system AGDL for attention-guided dual-layer image compression.
AGDL employs a CNN module to predict those pixels on and near a saliency sketch within ROI that are critical to perceptual quality.  Only the critical pixels are further sampled by compressive sensing. In addition, AGDL jointly decodes the two compression code layers for a much refined reconstruction, while strictly satisfying the transmitted CS constraints on perceptually critical pixels.

\vskip -0.2cm
\section*{Acknowledgments}
\vskip -0.1cm
This project is supported by Natural Sciences and Engineering Research Council of Canada (NSERC) and Huawei Canada.  The algorithms were implemented in part in the MindSpore framework.

{\small
\bibliographystyle{ieee_fullname}
\bibliography{AGDL}
}

\end{document}